\documentclass[letterpaper]{article} % DO NOT CHANGE THIS
\usepackage{aaai24}  % DO NOT CHANGE THIS
\usepackage{times}  % DO NOT CHANGE THIS
\usepackage{helvet}  % DO NOT CHANGE THIS
\usepackage{courier}  % DO NOT CHANGE THIS
\usepackage[hyphens]{url}  % DO NOT CHANGE THIS
\usepackage{graphicx} % DO NOT CHANGE THIS
\usepackage{natbib}  % DO NOT CHANGE THIS AND DO NOT ADD ANY OPTIONS TO IT
\usepackage{caption} % DO NOT CHANGE THIS AND DO NOT ADD ANY OPTIONS TO IT
\usepackage{algorithm}
\usepackage{algorithmic}

\usepackage{amsmath}
\usepackage{booktabs}
\usepackage{comment}
\usepackage{multirow}

\usepackage[capitalize]{cleveref}
\crefname{section}{Sec.}{Secs.}
\Crefname{section}{Section}{Sections}
\Crefname{table}{Table}{Tables}
\crefname{table}{Tab.}{Tabs.}

\usepackage{newfloat}
\usepackage{listings}
\DeclareCaptionStyle{ruled}{labelfont=normalfont,labelsep=colon,strut=off} % DO NOT CHANGE THIS
\floatstyle{ruled}
\newfloat{listing}{tb}{lst}{}
\floatname{listing}{Listing}
\title{OWQ: Outlier-Aware Weight Quantization \\for Efficient Fine-Tuning and Inference of Large Language Models}
\author{
    Changhun Lee\textsuperscript{\rm 1}\equalcontrib,
    Jungyu Jin\textsuperscript{\rm 2}\equalcontrib,
    Taesu Kim\textsuperscript{\rm 3},
    Hyungjun Kim\textsuperscript{\rm 3},
    Eunhyeok Park\textsuperscript{\rm 2}
}
\affiliations{
    \textsuperscript{\rm 1}Department of Convergence IT Engineering, POSTECH \\
    \textsuperscript{\rm 2}Graduate School of Artificial Intelligence, POSTECH \\
    \textsuperscript{\rm 3}SqueezeBits Inc. \\
    \{changhun.lee, jgjin0317\}@postech.ac.kr,
    \; \{taesu.kim, hyungjun.kim\}@squeezebits.com, \;eh.park@postech.ac.kr
}

\begin{document}

\maketitle

\begin{abstract}
Large language models (LLMs) with hundreds of billions of parameters require powerful server-grade GPUs for inference, limiting their practical deployment. To address this challenge, we introduce the outlier-aware weight quantization (OWQ) method, which aims to minimize LLM's footprint through low-precision representation. OWQ prioritizes a small subset of structured weights sensitive to quantization, storing them in high-precision, while applying highly tuned quantization to the remaining dense weights. This sensitivity-aware mixed-precision scheme reduces the quantization error notably, and extensive experiments demonstrate that 3.1-bit models using OWQ perform comparably to 4-bit models optimized by OPTQ. Furthermore, OWQ incorporates a parameter-efficient fine-tuning for task-specific adaptation, called weak column tuning (WCT), enabling accurate task-specific LLM adaptation with minimal memory overhead in the optimized format. OWQ represents a notable advancement in the flexibility, efficiency, and practicality of LLM optimization literature. The source code is available at https://github.com/xvyaward/owq.
\end{abstract}

\section{Introduction}
Large language models (LLMs)~\cite{brown2020language,radford2019language,scao2022bloom,touvron2023llama,zhang2022opt} demonstrate impressive generation performance on a wide range of complex language tasks, triggering explosive growth in LLM-based applications.
However, the extensive memory and computational demands are major obstacles to the widespread use of LLMs, not only for training but for inference as well. For instance, using fp16, the GPT3-175B model necessitates approximately 330 GB of space merely to store model parameters, which eventually costs hundreds of thousands of dollars to build the system with multiple server-grade GPUs. For the widespread adoption of LLMs, it is crucial to minimize such serving costs.

Recently, weight quantization has emerged as an attractive optimization method for LLMs~\cite{frantar2023optq}. By storing parameters into low-precision representation, storage space can be considerably saved, and this also introduces performance benefits by addressing memory bottlenecks and reduced communication costs~\cite{park2023lutgemm}. Advanced studies~\cite{frantar2023optq, park2023lutgemm} have shown that matrix multiplication with 3-bit weight and fp16 activation exhibits remarkable performance improvements on a single GPU compared to the case with fp16 weight and activation on multiple GPUs. Weight quantization of LLMs could resolve the memory and performance issues of LLMs jointly. 

Previously, OPTQ \cite{frantar2023optq}, also known as GPTQ~\cite{frantar2022gptq}, introduced a layer-wise post-training quantization (PTQ) method based on the optimal brain compression (OBC) algorithm~\cite{frantaroptimal}.
Notably, the compressed 3-bit OPT-175B model outperforms the fp16 OPT-30B model, even though they have similar memory footprints. 
The compressed model now occupies around 63 GB, allowing deployment on a single A100 GPU. However, there is room for improvement. OPT-175B still experiences some degradation with 3-bit quantization, and this effect is more pronounced in smaller models. Given that various model sizes are optimal for different scenarios, maintaining their accuracy remains highly important.
 
On the other hand, weight quantization broadens its application to task-specific fine-tuning. For example, QLoRA~\cite{qlora2023} enables the fine-tuning of the quantized LLMs for target tasks by incorporating low-rank high-precision tensors into the quantized dense matrix. During this process, only the added tensors are updated, while the quantized path remains unchanged. This method is gaining attention as it allows for efficient fine-tuning with reduced memory consumption while mitigating the drawback, quantization errors, during fine-tuning. However, the quality of the low-precision dense matrix is often overlooked, even though its suboptimal quality could diminish the benefits of fine-tuning. Even for the adaptation, high-quality weight quantization is essential.

\begin{figure*}
\centering
\includegraphics[width=\linewidth]{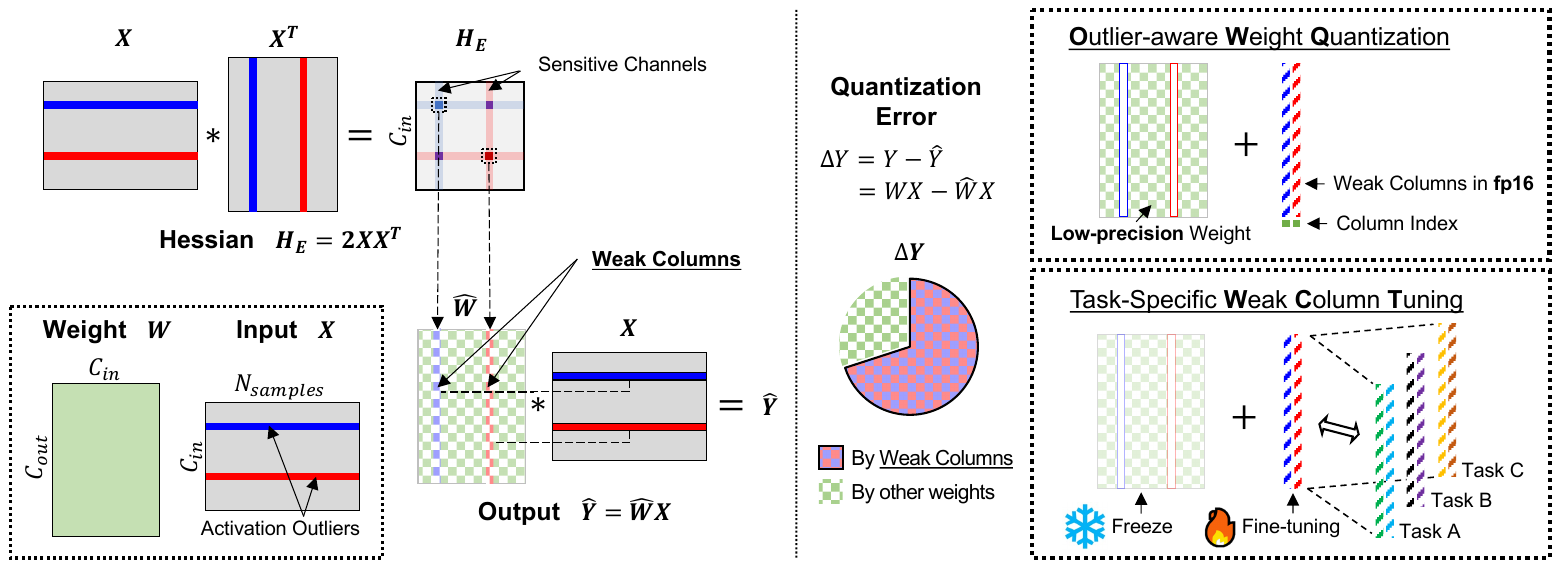}
\caption{The overview of the proposed weak columns concept, Outlier-aware Weight Quantization (OWQ) scheme, and Weak Column Tuning (WCT) scheme for efficient task-specific fine-tuning.}
\label{fig:owq_wct_main}
\end{figure*}

In this paper, we introduce a new weight quantization technique, Outlier-aware Weight Quantization (OWQ). OWQ is specifically designed considering the unique characteristics of LLMs, the presence of activation outliers~\cite{dettmers2022llm, weioutlier, xiao2022smoothquant}. Our analysis reveals that these outliers play a key role in the quality degradation of weight quantization. Based on this finding, we design OWQ that applies a mixed-precision quantization considering the sensitivity of each weight column. Extensive analysis indicates that the 3.1-bit OWQ model has comparable quality to the 4-bit OPTQ model. Moreover, we introduce an effective fine-tuning method based on OWQ, called Weak Column Tuning (WCT); it shares the benefit of OWQ during fine-tuning and inference, so the network can be updated to the target task with minimal memory overhead and accelerated during inference. 
Furthermore, the WCT-based fine-tuning model adapts with fewer trainable parameters than existing methods and outperforms them because of the superior representation quality of the weights quantized by OWQ.
To our knowledge, this is the first study to consider the existence of activation outliers in extremely low-precision weight quantization and closely integrate it with fine-tuning. 

\section{Background and Related Works} \label{sec:background}
\subsection{Quantization and LLMs}
Quantization is a widely used optimization technique aimed at exploiting the benefit of low-precision while maintaining the quality of the network. While its primary benefit is size reduction, quantization can also substantially accelerate performance through support for low-precision operations. However, a trade-off exists: quantization can lead to quality degradation, which numerous studies have aimed to address. Early studies focused on quantization-aware training (QAT) \cite{esser2019learned, zhou2016dorefa}, which tried to restore quality through additional training; however, as the understanding of quantization grew and various techniques emerged, post-training quantization (PTQ) \cite{librecq, nagel2019data, nagel2020up, weiqdrop} have been actively studied, enabling quality preservation without training. 

Due to the LLM's necessity of significant storage space and computational resources, it is crucial to apply optimization via quantization. In general, QAT has been favored for extremely low-bit precision to minimize quantization error. However, it is less favorable for LLM quantization because of the high cost of the training environment. Instead, PTQ has emerged as an important topic for LLM quantization. This field has two distinct approaches: one aims to quantize both activations and weights to int8~\cite{dettmers2022llm, xiao2022smoothquant}, considering both capacity reduction and acceleration. 
In contrast, the second approach focuses solely on weight quantization to sub-4-bit precision~\cite{frantar2023optq,park2023lutgemm}. In this paper, we align our work with the latter approach. While concentrating on weight quantization, we devise a novel quantization scheme, drawing significant inspiration from int8-related research on activation quantization.

\subsection{Int8 Quantization for Activation and Weight}

Int8 multiplication can provide up to 2x performance improvements and more than 5x energy consumption reduction compared to fp16 baselines \cite{horowitz20141}. Numerous studies~\cite{dettmers2022llm, xiao2022smoothquant} aim to quantize both activation and weight to int8 for matrix multiplication operations in LLMs. However, those studies identify a unique challenge of LLMs for activation quantization. LLMs exhibit a few outliers in intermediate activations, with values significantly larger than others, and these outliers are concentrated in specific feature dimensions. Preserving the values of these outliers is known to be crucial for maintaining accuracy after activation quantization.

In this study, while only weight quantization is applied, we figure out that the presence of activation outliers still impacts the sensitivity of weight quantization. We also demonstrate that considering activation outliers is essential for accurate weight quantization.

\subsection{OPTQ: Weight Quantization for LLMs}
OPTQ~\cite{frantar2023optq} is the state-of-the-art research in the field of weight quantization for LLMs. It is based on Optimal Brain Compression (OBC)~\cite{frantaroptimal}, which employs element-wise quantization (pruning) and compensation, using a Hessian-based metric of layer-wise quantization errors (\cref{eq: gptq} and \cref{eq: gptq2}). This approach differs from previous studies, which applied quantization through gradient descent based on the straight-through estimator~\cite{librecq,weiqdrop} or the rounding-to-nearest mechanism~\cite{dettmers2022llm}.

\begin{equation}
\label{eq: gptq}
    w_q = argmin_{w_q} \frac{(quant(w_q)-w_q)^2}{[H_F^{-1}]_{qq}},
\end{equation}
\begin{equation}
\label{eq: gptq2}
    \delta _F = - \frac{w_q - quant(w_q)}{[H_F^{-1}]_{qq}} \cdot (H_F^{-1})_{:,q}
\end{equation}

OPTQ has optimized OBC to parallelize quantization for each element of the output channel dimension, enabling rapid quantization. While it showcases the potential of sub-4-bit quantization for LLMs, reducing the model size or increasing the problem's complexity results in decreased accuracy compared to the fp16 baselines. In this paper, we suggest selectively applying high-precision to weights that are vulnerable to quantization caused by activation outliers and applying the OPTQ to the remaining weights with the modification based on quantization configuration tuning for additional error reduction. These enhancements can significantly reduce the quantization error while preserving the quantization speed of OPTQ.

\subsection{Parameter-Efficient Fine-Tuning (PEFT)}
Fine-tuning LLMs for specific tasks can improve performance on unseen or complex tasks, but the large number of parameters requires a hyperscale computation system, which is often impractical due to high costs. Therefore, Parameter-Efficient Fine-Tuning (PEFT) schemes have been introduced to address this issue. LoRA~\cite{lora2022} exemplifies PEFT by freezing pre-trained weights but incorporating a small fraction of learnable parameters through low-rank decomposition. As only the added parameters are updated, LoRA can be adapted using significantly less memory. QLoRA~\cite{qlora2023} further reduces memory consumption by replacing dense weights with quantized weights, making the fine-tuning process more lightweight. Since fine-tuning can mitigate quantization errors, QLoRA emerges as a desirable optimization for the practical, task-specific deployment of LLMs. In this study, we introduce an OWQ-compatible PEFT scheme. The superior representation quality of OWQ yields exceptional performance after PEFT with lower resource overhead than QLoRA.

\section{Problem Definition and Motivation}

In this section, before introducing our idea, we first define the problem and explain our findings clearly. The proposed OWQ is designed to apply layer-wise uniform quantization for the weights of LLMs with minimal quality degradation. When an input feature $X \in R^{C_{in} \times N}$ is given, where $C_{in}$ represents the number of input channels and $N$ is the sequence length of the input, the full-precision weight matrix $W \in R^{C_{out} \times C_{in}}$ for $C_{out}$ output features are mapped to low-precision toward minimizing the difference of output activations before and after quantization. The objective function to find the quantized weight $\hat{W}$ that minimizes the squared error is defined as follows:
\begin{equation}
\underset{{\hat{W}}}{\arg\min} \; E = \underset{{\hat{W}}}{\arg\min} \; || W X - \hat{W} X ||_2^2 \quad \text{s.t.} \quad C(\hat{W})<C_t,
\label{eq:error}
\end{equation}
where $C(\cdot)$ represents the compression ratio and $C_t$ is the target compression ratio. 
The layer-wise quantization process is applied sequentially from the model input to the output, ensuring the quantization of all weights in the model. We keep embedding and head weights with full-precision.

\subsection{Layer-wise Quantization and Hessian of Weights}
In this subsection, we explain our insight concerning the relationship between weight sensitivity and activation outliers in the context of quantization: weights linked to activation outliers are particularly susceptible to quantization. This understanding serves as the core motivation of OWQ.

Initially, we restructure the squared error term in \cref{eq:error} to represent the sum of squared errors for each output channel within the weight matrix, resulting in the equation $\Sigma_{i=1}^{C_{out}} ||W_{i,:} X - \hat{W}_{i,:} X||^2_2$. This decomposition distinctly showcases that the overall error is divided into individual errors for each output channel. With the modified equation, our focus shifts to two key aspects. Firstly, it is important to note that there is no Hessian interaction between output channels. Specifically, the individual Hessians with respect to the layer-wise quantization error, denoted as $H^{(i)} \in R^{C_{in} \times C_{in}}$, have an identical value as:
\begin{equation}
H^{(i)} = H = \frac{\partial^2E_{i}}{\partial W_{i,:}^2} = 2XX^T.
\label{eq:hessian}
\end{equation}

Secondly, as observed in previous studies~\cite{nagel2020up}, the individual error term can be approximated using Taylor expansion. By setting $\Delta W_{i,:} = W_{i,:} - \hat{W}_{i,:}$, the error for the $i$-th output channel can be expressed as follows:
\begin{equation}
E_i = ||W_{i,:}X - \hat{W}_{i,:}X||_2^2 \approx \Delta W_{i,:} H \Delta W_{i,:}^T.
% E_i = ||W_{i,:}X - \hat{W}_{i,:}X||_2^2 = \Delta W_{i,:} H_i \Delta W_{i,:}^T.
\label{eq:error_H}
\end{equation}
For detailed proof, please refer to the appendix. The equation shows that in the context of layer-wise quantization, the output error can be directly related to the Hessian and the magnitude of weight perturbation.

Keeping these observations in mind, we can derive an interesting insight by acknowledging the presence of activation outliers in LLMs. Previous studies~\cite{dettmers2022llm, xiao2022smoothquant} have reported that certain feature dimensions of LLM activation contain outliers with significantly larger values than others. As shown in \Cref{fig:owq_wct_main} Left, these activation outliers make some elements of $H$ have exceptionally high values. This abnormal surge in Hessian values increases the corresponding weight channels' sensitivity to quantization. 
In details, as indicated in \cref{eq:error_H}, even when the same degree of weight perturbation is present, the ensuing change in output can be considerably larger due to some large elements of $H$.
We refer to the weights susceptible to quantization as weak column, specifically those associated with the activation outliers in a specific input channel.

\begin{figure}
\centering
\includegraphics[width=0.8\columnwidth]{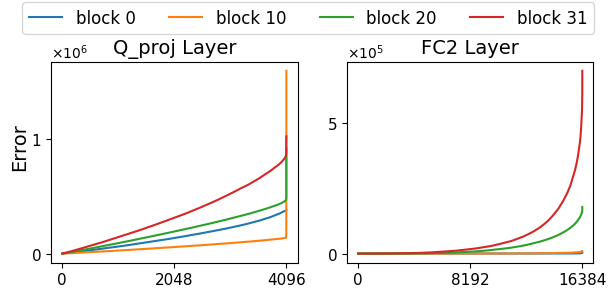}
\caption{Cumulative error for each input channel. The x-axis channel indices are sorted in ascending order based on their influence on the output error.}
\label{fig:3bit_cdf}
\end{figure}

Therefore, if we quantize all weights to the same bit-width during the weight quantization process, the quantization error from the weak columns corresponding to the activation outliers can lead to substantial perturbation on the output, resulting in a notable quantization error. \Cref{fig:3bit_cdf} supports this assertion, indicating that a large portion of the error originates from a limited number of channels, which align with the weak columns. To minimize the error from weak columns, it's imperative to specially address these columns.

\section{OWQ: Outlier-aware Weight Quantization}\label{method}
To tackle this issue, we introduce a novel concept termed Outlier-aware Weight Quantization (OWQ). OWQ encompasses two steps: initially, it identifies the weak columns and excludes them from quantization. Subsequently, it quantizes the remaining weights to extreme low-precision using meticulously tuned quantization parameters. The overview of the proposed OWQ scheme is illustrated in \Cref{fig:owq_wct_main} Left. In this section, we thoroughly discuss the details of the OWQ implementation.

In ~\cref{eq:hessian}, we highlighted the relationship between the Hessian matrix and sensitivity caused by activation outliers. We also demonstrated that the final error is influenced by the quadratic terms of perturbations with the Hessian matrix in ~\cref{eq:error_H}. Building on these insights, we define the sensitivity of $j$-th weight column as follows: 
\begin{equation}
    sensitivity_j = \lambda_j || \Delta W_{:,j} ||_2^2,
\label{eq: sensitivity}
\end{equation}
where $\lambda_j$ is the $j$-th diagonal element of the Hessian matrix. By analyzing the sensitivity of individual columns, we can effectively identify the weak columns that are vulnerable to quantization and require higher precision. When the goal is to select a specific number ($k$) of weak columns, the proposed metric is utilized to choose the top-$k$ sensitive columns based on their sensitivity values.

It is worth noting that hessian-based metrics have been often used in previous studies~\cite{dong2019hawq,dong2020hawq}. While our work has a distinctive difference in the granularity of the quantization domain and detailed expression, our observation is well aligned with the intuition of the existing studies. 

Following the selection of weak columns, the remaining weights are quantized into low-precision. Any low-precision quantization scheme is applicable, but we employ OPTQ. Since OPTQ also utilizes sequential column-wise quantization, the weights excluding the weak columns can be seamlessly integrated into the OPTQ framework. Additionally, please note that the weak columns can be used to further mitigate errors that occur during the OPTQ process. OPTQ updates the remaining unquantized weights to compensate for the errors caused by quantizing the weight columns in the current step, as shown in \cref{eq: gptq2}. By rearranging the high-precision weak columns to the end of the weight before utilizing OPTQ, quantization errors from other columns during the OPTQ process can be largely compensated for by the weak columns. Since the weak columns are retained with full-precision, these compensated values can be preserved even if all other columns are quantized.

After identifying the weak columns and quantizing the remaining weights, we store the weak columns as fp16 and use an extra single integer per column, which represents the column index of the weak columns. In addition, we store a low-precision matrix whose positions of weak columns are zero-filled. Therefore, compared to OPTQ, the additional storage overhead is solely caused by the weak columns. This overhead is negligible ($\approx$ 0.3\%), while the accuracy is significantly improved. In addition, we also provide the specialized acceleration for OWQ format on real GPU. A comprehensive explanation of the implementation will be provided in the ``Acceleration on Real Device" section.

\begin{figure}
\centering
\includegraphics[width=0.95\columnwidth]{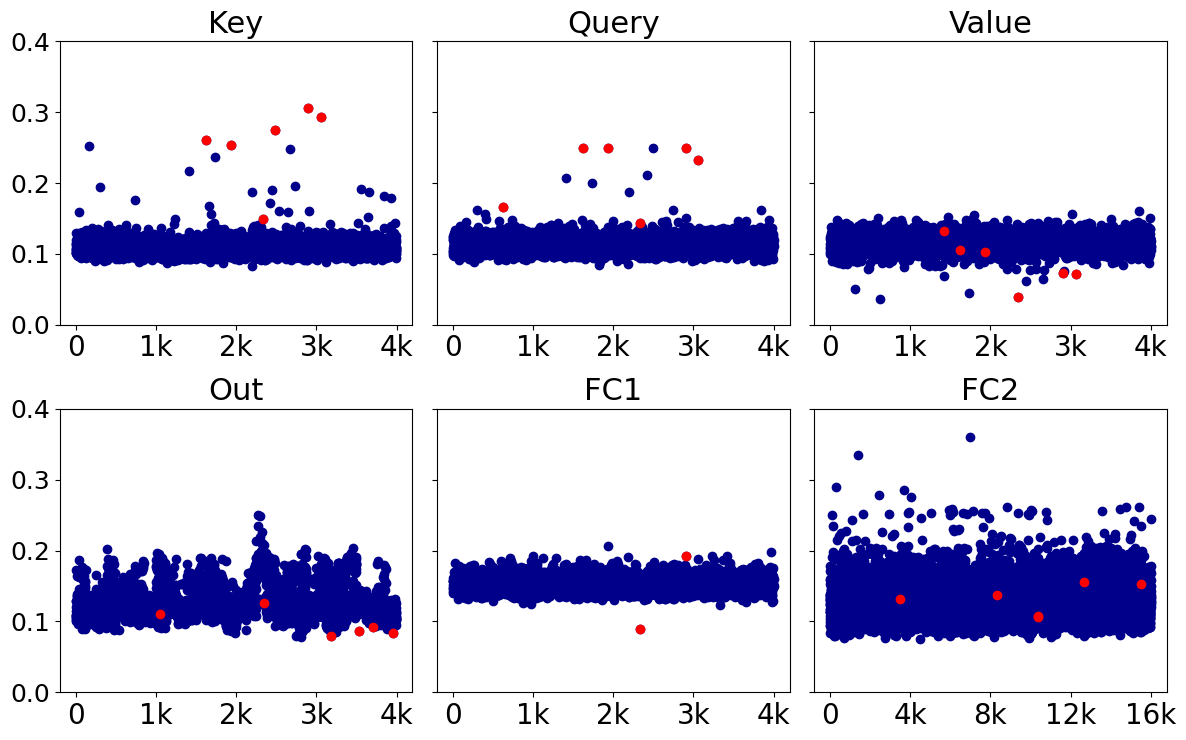}
\caption{Min-max range of the weights in each column (blue dots) and the selected weak columns (red dots).}
\label{fig:w_outlier}
\end{figure}

Additionally, as depicted in \Cref{fig:w_outlier}, our approach is distinctly different from existing outlier-aware quantization studies~\cite{park2018energy,park2018value}. While the naming is similar, the outcome is completely different. In those studies, outlier weights are excluded from quantization based solely on their magnitude to minimize the error of weights before and after the quantization. However, our method minimizes the error of output activation. As shown in the figure, the weights are selected based on their sensitivity rather than their magnitude.

\subsection{Quantization Configuration Search}

We mentioned that we use OPTQ for the quantization of the dense weights except for the weak columns. However, we made an important modification to OPTQ to further reduce quantization error. While OPTQ originally relies on a straightforward min-max quantization approach, we have incorporated the benefits of truncation. Many previous studies~\cite{esser2019learned, librecq, nahshan2021loss, weiqdrop} pointed out that the truncation reduces the overall quantization error significantly, improving the quality of output significantly. In OWQ, we searched the quantization configurations, such as step size and zero point using a simple 2D grid search, to narrow the quantization range. The optimal values of the parameters that minimize the difference before and after quantization are searched via rounding to the nearest with truncation, and we apply OPTQ on top of the searched values. 

This modification substantially reduces the overall reconstruction error, significantly improving quality.
In OWQ, the WikiText-2 perplexity is lowered from 12.14 to 11.21 for OPT-6.7B.
An interesting observation is that when truncation is applied to conventional OPTQ, output quality degrades, with the WikiText-2 perplexity increasing from 12.88 to 48.26 in OPT-6.7B. We have observed that the weak columns in the key and query layers of transformer blocks have exceptionally large values. Truncating these leads to substantial error, resulting in a significant reduction in accuracy. However, with our approach, this error is avoided because weak columns are retained as full-precision. We argue that this enhancement is a significant advantage of OWQ, notably improving the representation quality of the low-precision matrix.

\subsection{PEFT with Weak Column Tuning}
When applied to pre-trained LLMs, OWQ results in minimal quality degradation, allowing the model to maintain its zero-shot generative performance. Nonetheless, to further boost output quality, fine-tuning might be necessary for newly introduced tasks. In this context, we present a Weak Column Tuning (WCT) scheme, illustrated in \Cref{fig:owq_wct_main} Right. 
WCT first quantizes the base model with OWQ and then fine-tunes only the weak columns that retained high-precision as a result of OWQ. 
In OWQ, we identify weak columns using channel-wise sensitivity and retain them in high-precision, meaning slight alterations in these sensitive columns have a substantial impact on the output. Moreover, because they are high-precision (fp16), the values can be freely modified. Empirically, we've found that updating these weak columns facilitates the adaptation of the quantized network to the desired task. It's worth noting that the low-precision dense matrix remains static during this update. 

Because learnable weak columns represent only a small fraction of the overall parameters, updates also result in negligible memory overhead. Furthermore, the OWQ format is preserved after fine-tuning, enabling us to leverage the benefits of acceleration with the customized kernel for OWQ. Additionally, by allocating per-task weak columns, we can serve multiple task-specific models with minimal memory overhead through shared utilization of the dense low-precision matrix, which constitutes the majority of the memory footprint.

The versatile adaptability of OWQ broadens its range of applications. Furthermore, as will be shown in the ``Experiment" section, WCT surpasses leading fine-tuning methods in terms of both memory usage and output quality. This is because OWQ improves the quality of dense low-precision representation, and sensitive columns are pre-picked and used as the update target. These factors enhance the quality of the fine-tuned network significantly. 
Please refer to the ``Comparison of PTQ Methods used in WCT" section for details.

\begin{table*}[t]
    \begin{center}
    \small
    \begin{tabular}{ c | c | l l l l l l l l }
        \toprule
        OPT & Bits & \quad 125M & 350M & 1.3B & 2.7B & 6.7B & 13B & 30B & 66B   \\
        \midrule
        full & 16 & \quad 27.65 & 22.00 & 14.63 & 12.47 & 10.86 & 10.13 & 9.56 & 9.34 \\
        \midrule
        RTN & 4 & \quad 37.28 & 25.94 & 48.17 & 16.92 & 12.10 & 11.32 & 10.98 & 110  \\
        OPTQ & 4 & \quad 32.05 & 23.87 & 15.47 & 12.83 & 11.14 & 10.29 & 9.57 & 9.34  \\
        \textbf{OWQ} & 4.01 & \quad $\textbf{29.47}^{*}_{\pm.17}$ & $\textbf{23.19}^{*}_{\pm.16}$ & $\textbf{15.01}_{\pm.06}$ & $\textbf{12.39}_{\pm.02}$ & $\textbf{10.87}_{\pm.02}$ & $\textbf{10.26}_{\pm.02}$ & $\textbf{9.50}_{\pm.01}$ & $\textbf{9.25}_{\pm.04}$ \\
        \midrule
        OPTQ & 3 & \quad 53.43 & 32.28 & 20.90 & 16.55 & 12.88 & 11.58 & 10.29 & 9.90  \\
        \textbf{OWQ} & 3.01 & \quad $\textbf{35.26}^{*}_{\pm.61}$ & $\textbf{26.59}^{*}_{\pm.51}$ & $\textbf{16.40}_{\pm.15}$ & $\textbf{13.21}_{\pm.09}$ & $\textbf{11.21}_{\pm.05}$ & $\textbf{11.48}_{\pm.03}$ & $\textbf{9.61}_{\pm.02}$ & $\textbf{9.28}_{\pm.03}$\\
        \textbf{OWQ} +TS+AO & 3.01 & \quad $\textbf{35.19}^{*}_{\pm.68}$ & $\textbf{25.86}^{*}_{\pm.25}$ & $\textbf{15.96}_{\pm.18}$ & $\textbf{13.28}_{\pm.06}$ & $\textbf{11.20}_{\pm.05}$ & $\textbf{11.14}_{\pm.02}$ & $\textbf{9.66}_{\pm.06}$ & $\textbf{9.31}_{\pm.01}$\\
        \textbf{OWQ} & 3.1 & \quad $\textbf{33.41}_{\pm.25}$ & $\textbf{26.00}_{\pm.14}$ & $\textbf{15.39}_{\pm.06}$ & $\textbf{12.98}_{\pm.10}$ & $\textbf{11.14}_{\pm.03}$ & $\textbf{10.38}_{\pm.01}$ & $\textbf{9.57}_{\pm.02}$ & $\textbf{9.30}_{\pm.02}$ \\
        \bottomrule
    \end{tabular}
    \end{center}
    \caption{OPT WikiText-2 perplexity ($\downarrow$). For the results with *, we used an extra 0.05 bits instead of 0.01 bits; there are few or no weak columns in the budget of 0.01 bits due to the small model dimension. TS: True-sequential and AO: Act-order options.}
    \label{table:opt_wiki}
\end{table*}

\begin{table}[t]
        \centering
        \small
        \begin{tabular}{@{\hskip 0.08in} c @{\hskip 0.08in} | @{\hskip 0.08in} c @{\hskip 0.08in} | @{\hskip 0.15in} l @{\hskip 0.1in} l @{\hskip 0.1in} l @{\hskip 0.1in} l }
            \toprule
            LLaMA & Bits & 7B & 13B & 30B & 65B \\
            \midrule
            full & 16 & 5.68 & 5.09 & 4.10 & 3.53 \\
            \midrule
            RTN & 4 & 6.29 & 5.53 & 4.54 & 3.92 \\
            OPTQ & 4 & 6.10 & 5.36 & 4.45 & 4.10 \\
            \textbf{OWQ} & 4.01 & $\textbf{5.94}_{\pm.02}$ & $\textbf{5.25}_{\pm.00}$ & $\textbf{4.25}_{\pm.00}$ & $\textbf{3.74}_{\pm.02}$ \\
            \midrule
            OPTQ & 3 & 8.13 & 6.67 & 5.67 & 5.41 \\
            \textbf{OWQ} & 3.01 & $\textbf{6.66}_{\pm.04}$ & $\textbf{5.66}_{\pm.01}$ & $\textbf{4.75}_{\pm.02}$ & $\textbf{4.25}_{\pm.01}$ \\
            \textbf{OWQ} & 3.1 & $\textbf{6.41}_{\pm.01}$ & $\textbf{5.56}_{\pm.01}$ & $\textbf{4.63}_{\pm.01}$ & $\textbf{4.09}_{\pm.01}$ \\
            \bottomrule
        \end{tabular}
        \caption{LLaMA WikiText-2 perplexity (lower is better).}
        \label{table:llama_wiki}
\end{table}

\section{Experiments}
\label{experiments}
\subsection{Experimental Setup}
\label{experimental_setup}
To validate the outstanding performance of our proposed method, we present quantization results for large-scale LLMs such as OPT~\cite{zhang2022opt} and LLaMA~\cite{touvron2023llama} families. Our primary baseline is OPTQ, so we apply identical experimental settings of it. For instance, our calibration dataset consists of 128 random 2048 token segments from the C4 dataset~\cite{raffel2020exploring}. Experiments were conducted on a single NVIDIA A100 GPU with 80 GB of main memory or RTX3090 GPU with 24 GB memory. Like OPTQ, our method quantizes the target model without re-training.
To measure the zero-shot or few-shot performance, we utilize an open-source evaluation repository, EleutherAI/lm-evaluation-harness~\cite{eval-harness}. 
Please note that we report the numbers with an error margin based on 3 experiments with different seeds.

Compared to the baselines of 3-bit and 4-bit OPTQ results, we present two variants, with an extra 0.01 bit and 0.1 bit overhead, respectively. For the 4-bit baseline, we found that 0.01 bit overhead is sufficient. The additional storage area of extra bit is evenly distributed across the linear layers within the transformer block. For instance, in the OPT model which has six linear layers (key, query, value, out, fc1, and fc2), the weak columns of the key layers contribute an average of 0.00167 bit in the 3.01-bit configuration (0.15\% columns of the key weight matrix). This extra bit covers the all overhead of the mixed-precision representation. If we quantize OPT-175B model with an average of 3.01 bits, it will require approximately 260 MB of additional storage compared to the 3-bit OPT-175B OPTQ model, which utilizes around 63.1 GB of storage space. All experiments were conducted using the PyTorch 2.0 \cite{paszke2019pytorch} framework with HuggingFace integration~\cite{wolf2019huggingface}.

While not discussed in OPTQ papers, the ``act-order" (AO) option was recently added to their official GitHub. This option quantizes columns in order according to the activation magnitude and is distinct from the earlier OPTQ approach that quantized columns sequentially. Similarly, the ``true-sequential" (TS) option was introduced, which applies quantization sequentially, taking into account the quantization error from previous layers within a block. Both methods boost the accuracy of OPTQ, and for a comprehensive comparison, we provide OPTQ results with ``TS + AO" configuration by default. We don't use these options for OWQ unless otherwise noted.

\begin{table*}[t]
    \begin{center}
    \small
    \begin{tabular}{ c | c | l l l l l l l l  }
        \toprule
        OPT & Bits & \quad 125M & 350M & 1.3B & 2.7B & 6.7B & 13B & 30B & 66B   \\
        \midrule
        full & 16 & \quad 26.78 & 28.77 & 36.34 & 40.27 & 44.22 & 45.34 & 47.97 & 49.86  \\
        \midrule
        \textbf{OWQ} & 4.01 & \quad $\textbf{26.67}^{*}_{\pm.29}$ & $\textbf{28.85}^{*}_{\pm.11}$ & $\textbf{36.48}_{\pm.19}$ & $\textbf{39.75}_{\pm.05}$ & $\textbf{43.94}_{\pm.03}$ & $\textbf{45.42}_{\pm.14}$ & $\textbf{47.83}_{\pm.22}$ & $\textbf{49.57}_{\pm.07}$  \\
        OPTQ & 4 & \quad 26.43 & 28.31 & 36.41 & 39.66 & 43.69 & 44.97 & 47.58 & 49.47  \\
        \midrule
        \textbf{OWQ} & 3.1 & \quad $25.99_{\pm.28}$ & $27.77_{\pm.23}$ & $\textbf{35.86}_{\pm.13}$ & $\textbf{38.93}_{\pm.19}$ & $\textbf{42.80}_{\pm.07}$ & $\textbf{44.50}_{\pm.22}$ & $\textbf{47.20}_{\pm.28}$ & $\textbf{49.02}_{\pm.25}$ \\
        \textbf{OWQ} & 3.01 & \quad $\textbf{26.18}^{*}_{\pm.40}$ & $\textbf{27.91}^{*}_{\pm.17}$ & $\textbf{35.51}_{\pm.24}$ & $\textbf{38.79}_{\pm.81}$ & $\textbf{42.51}_{\pm.26}$ & $\textbf{44.62}_{\pm.13}$ & $\textbf{47.17}_{\pm.20}$ & $\textbf{48.84}_{\pm.03}$  \\
        OPTQ & 3 & \quad 26.00 & 27.78 & 34.01 & 37.16 & 41.63 & 43.70 & 46.65 & 48.01 \\
        \bottomrule
    \end{tabular}
    \end{center}
    \caption{The average value of three few-shot scores (\%): ARC-challenge,
Hellaswag, and MMLU for OPT families
(higher is better). For the results with *, we used an extra 0.05 bits instead of 0.01 bits due to the small model dimension.}
    \label{table:opt_lambada_openai}
\end{table*}

\begin{table}[t]
        \centering
        \small
        \begin{tabular}{@{\hskip 0.05in} c @{\hskip 0.05in} | @{\hskip 0.05in} c @{\hskip 0.05in} | @{\hskip 0.1in} l @{\hskip 0.08in} l @{\hskip 0.08in} l @{\hskip 0.08in} l @{\hskip 0.05in}}
            \toprule
            LLaMA & Bits & 7B & 13B & 30B & 65B \\
            \midrule
            full & 16 & 54.81 & 61.56 & 68.21 & 71.17 \\
            \midrule
            \textbf{OWQ} & 4.01 & $\textbf{54.01}_{\pm.04}$ & $\textbf{60.54}_{\pm.52}$ & $\textbf{68.12}_{\pm.15}$ & $\textbf{70.30}_{\pm.06}$ \\
            OPTQ & 4 & 52.91 & 59.90 & 66.96 & 69.92 \\
            \midrule
            \textbf{OWQ} & 3.1 & $\textbf{51.93}_{\pm.48}$ & $\textbf{58.84}_{\pm.36}$ & $\textbf{66.18}_{\pm.11}$ & $\textbf{69.26}_{\pm.29}$ \\
            \textbf{OWQ} & 3.01 & $\textbf{50.21}_{\pm.45}$ & $\textbf{57.69}_{\pm.21}$ & $\textbf{66.21}_{\pm.25}$ & $\textbf{68.82}_{\pm.29}$ \\
            OPTQ & 3 & 47.06 & 52.87 & 61.15 & 65.75 \\
            \bottomrule
        \end{tabular}
        \caption{The average value of three scores (\%): ARC-challenge,
Hellaswag, and MMLU for LLaMA families.
}
        \label{table:llama_zeroshot}
\end{table}

\subsection{Results of Perplexity Measure}\label{sec:perplexity}

The accuracy of the proposed model is assessed through the evaluation on multiple language tasks, including WikiText-2 \cite{merity2016pointer}, Penn Treebank (PTB) \cite{marcus1994penn}, and C4. Perplexity-based tasks are particularly sensitive to model quantization~\cite{frantar2023optq}, with perplexity numbers serving as indicators of the generative performance of the quantized model. The results for WikiText-2 can be found in \Cref{table:opt_wiki} and \Cref{table:llama_wiki}, while the results for PTB and C4 are provided in the appendix.

The results clearly demonstrate that OWQ consistently delivers substantial quality improvements across the LLM families, irrespective of the model size. The 3.01-bit OWQ model effectively mitigates the quality degradation observed in the 3-bit OPTQ model, while the 3.1-bit model achieves comparable performance to the 4-bit OPTQ model. Furthermore, OWQ 4.01-bit yields noteworthy improvements, highlighting the significance of treating weak columns. These results underscore the importance and effectiveness of our approach in preserving model quality after quantization. 

An interesting finding is the significant improvement in model quality for OPT models with less than 13 billion parameters when applying OWQ. Although previous studies have highlighted the presence of activation outliers in models with more than 6.7 billion parameters~\cite{dettmers2022llm}, even smaller models with moderately large channels can still benefit from mixed precision quantization. This suggests that the concept of weak columns remains valid and effective in enhancing the quality of LLMs, regardless of the model size.

In addition, please note that OWQ consistently outperforms OPTQ with TS + AO, but TS + AO doesn't give performance benefits to OWQ (\Cref{table:opt_wiki}).
Although no theoretical interpretation for those options was proposed, our study suggests that the benefits of ``act-order" arise from sensitivity-aware quantization. This means that applying sequential quantization beginning with sensitive columns improves performance for OPTQ. However, act-order alone cannot sufficiently mitigate the quality degradation caused by weak columns within a low-precision domain. Interestingly, the benefit of TS is also mitigated as quantization error is already greatly reduced by using OWQ.

\subsection{Results of Various Few-shot Tasks}
We conducted additional experiments on diverse few-shot language tasks. Referring to the “Open LLM Leaderboard” \cite{open-llm-leaderboard} from the Huggingface H4 team as a benchmark, we relied on the average scores from ARC-challenge (25-shot) \cite{clark2018think}, Hellaswag (10-shot) \cite{zellers2019hellaswag}, and MMLU (5-shot) \cite{hendrycks2021measuring}. In particular, MMLU is a collection of 57 tasks. These benchmarks were chosen because they test a range of reasoning and general knowledge across diverse fields, in few-shot contexts.
The average scores of \Cref{table:opt_lambada_openai} and \Cref{table:llama_zeroshot} convince us that the proposed OWQ is consistently superior to OPTQ in various model sizes, from 125m to 66B. In particular, we can see the considerable gap between OPTQ 3-bit and OWQ 3.01-bit for the LLaMA families in \Cref{table:llama_zeroshot} despite using the TS and AO options for OPTQ. Our method's strength lies in its universality, consistently boosting the performance of generative models with minimal storage overhead.

\subsection{Acceleration on Real Device}\label{sec:accel}
To demonstrate the advantages of low-precision acceleration, we developed a customized CUDA kernel for OWQ and assessed its latency overhead on an A100 GPU. First, we decompressed the low-precision matrix into the fp16 format and performed dense GeMV multiplication. At this stage, the overhead is identical to that of OPTQ's customized kernel. Additionally, we select the activation input channels for these weak columns on-the-fly and use another GeMV kernel specifically for them. This approach effectively avoids the issues of irregular memory access. \Cref{table:kernel_overhead} displays the kernel overhead of OWQ 3.01-bit for various model sizes. The mixed-process computation adds up to only 3.21\% latency compared to the 3-bit acceleration of the OPTQ kernel on the LLaMA 7B model, and the overhead is generally amortized for larger models.

\begin{table}
        \centering        
        \small
        \begin{tabular}{ c | c @{\hskip 0.12in} c @{\hskip 0.12in} c | c | c @{\hskip 0.12in} c @{\hskip 0.12in} c }
            \toprule
            OPT & 6.7B & 13B & 66B & LLaMA & 7B & 13B & 65B \\
            \midrule
            k/q/v/o & 3.07 & 3.11 & 2.84 & k/q/v/o & 3.21 & 3.47 & 2.34 \\
            fc1 & 2.31 & 1.53 & 1.65 & up/gate & 2.34 & 2.01 & 1.65 \\
            fc2 & 2.15 & 2.04 & 2.23 & down & 2.03 & 2.21 & 2.03 \\
            \bottomrule
        \end{tabular}
        \caption{Kernel overhead of OWQ 3.01-bit (vs OPTQ) (\%).}
        \label{table:kernel_overhead}
\end{table}

\subsection{Quantization Speed}

For LLM quantization, the quantization algorithm's speed is crucial. While OWQ adds operations for weak column selection and hyperparameter tuning compared to OPTQ, sharing the Hessian with OPTQ minimizes OWQ's overhead. Furthermore, applying the true-sequential option to OPTQ adds extra runtime for OPTQ, further reducing the gap with OWQ quantization time. On A100 GPU, OWQ can quantize a 66B model in under 3 hours, presenting its practicality.

\subsection{Results of WCT-based Fine-tuning}

\begin{figure}[t]
     \centering
     \includegraphics[width=0.95\columnwidth]{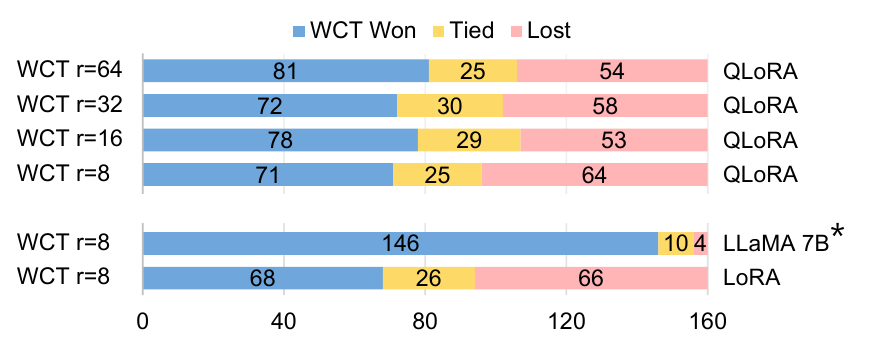}
     \caption{GPT-4 based analysis of WCT-based fine-tuning. * denotes the base model (without fine-tuning).}
     \label{fig:finetune}
\end{figure}

To validate the superior performance of Weak Column Tuning (WCT) for task-specific adaptation, we fine-tuned the LLaMA 7B model with 4-bit quantization using WCT and then compared the results. To obtain the results of QLoRA \cite{qlora2023}, we used the official checkpoint. For a fair comparison, we used the same subset of the OpenAssistant \cite{kopf2023openassistant} dataset that QLoRA employed. We evaluated performance by inputting the results generated by both models for 80 questions from the Vicuna Benchmark \cite{vicuna2023} into GPT-4 \cite{openai2023gpt4} to determine which was better. As reported in a previous work \cite{qlora2023}, we found a bias in which GPT-4 favors the system that appears first in a given pair of systems, giving it a higher score. To eliminate this bias, we evaluate both ordering cases and report results for a total number of 160 evaluations. LoRA \cite{lora2022} and QLoRA both use the 64 rank of adapter modules. We used nucleus sampling with p = 0.9 and temperature 0.7 for all generations.

In the WCT experiments, $r=k$ means the configuration with r weak columns for each layer. As depicted in \Cref{fig:finetune}, WCT with 64 weak columns ($r=64$) surpasses the QLoRA. In other words, GPT-4 evaluated the tuning results using WCT as better more often (81 vs 54 for $r=64$ case). 
Remarkably, WCT with just 8 weak columns (takes only 6.8\% of learnable parameters vs. QLoRA) outperforms QLoRA and yields results comparable to full-precision LoRA.
As the quality of the compressed weight is on par with the full-precision model, OWQ combined with WCT delivers performance matching that of full-precision LoRA only with 24.4\% of overall memory usage during inference. WCT only updates weak columns, which are highly sensitive to update. This feature makes WCT compensate accuracy with a smaller rank than the conventional LoRA. With OWQ + WCT, we can enjoy the benefits of quantization not only in inference but also in task-specific adaptation.

\begin{figure}[t]
     \centering
     \includegraphics[width=\columnwidth]{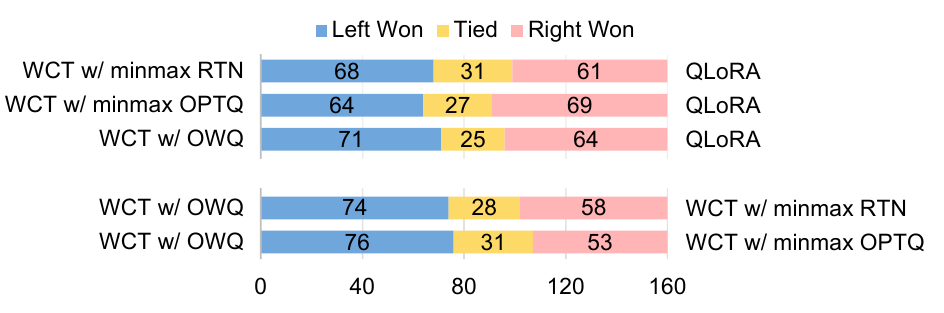}
     \caption{Comparison of fine-tuned performance of several post-training quantization methods used for WCT.}
     \label{fig:finetune_ptq}
\end{figure}

\subsection{Comparison of PTQ Methods used in WCT}
We compare several quantization methods used for quantizing fixed dense weights in WCT and verify that the sophisticated quantization method is important for performance after fine-tuning. After quantizing the LLaMA-7B model with different quantization methods, the models refined with WCT are asked to generate answers to questions from the Vicuna Benchmark, and the quality of the answers is compared using GPT-4-based evaluation. 
For all quantization methods, we used $r=8$ as the number of weak columns and linear asymmetric quantization with the per-channel granularity.
Although the performance loss due to quantization can be compensated by fine-tuning, \Cref{fig:finetune_ptq} shows that the quality of the quantized model before fine-tuning affects the performance after fine-tuning.

\begin{table}[t]
        \centering
        \small
\begin{tabular}{lcccc}
\toprule
\multirow{2}{*}{Method / Bits / Group} & \multicolumn{4}{c}{OPT-6.7B}\\ \cline{2-5}
 & Wiki2 & PTB & C4 & lamb. $\uparrow$   \\ 
\midrule
OPTQ / 3 & 12.76 & 19.35 & 14.55 & 60.15 \\
OPTQ / 3 \quad \, / g1024 & 12.05 & 18.01 & 13.89 & 64.44   \\
OPTQ / 3 \quad \, / g128 & 11.55 & 17.26 & 13.43 & 64.96  \\ 
\midrule
\textbf{OWQ} / 3.01 & 11.22 & 16.32 & 13.23 & 69.94   \\
\textbf{OWQ} / 3.01 / g1024 & 11.18 & 16.32 & 13.19 & 68.91   \\
\textbf{OWQ} / 3.01 / g128 & 11.16 & 16.23 & 13.10 & 68.61 \\ \bottomrule
 
 \end{tabular}
 \begin{tabular}{lcccc}
 \toprule
\multirow{2}{*}{Method / Bits / Group}  & \multicolumn{4}{c}{LLaMA 7B} \\ \cline{2-5}
 & Wiki2 & PTB & C4 & winog. $\uparrow$ \\ 
\midrule
OPTQ / 3 & 8.08 & 14.13 & 10.26 & 64.33 \\
OPTQ / 3 \quad \, / g1024 & 7.15 & 12.75 & 9.12 & 66.26 \\
OPTQ / 3 \quad \, / g128 & 6.56 & 12.48 & 8.36 & 67.60 \\ 
\midrule
\textbf{OWQ} / 3.01 & 6.65 & 12.47 & 8.62 & 67.05 \\
\textbf{OWQ} / 3.01 / g1024 & 6.61 & 12.05 & 8.49 & 67.60 \\
\textbf{OWQ} / 3.01 / g128 & 6.40 & 11.72 & 8.18 & 69.57 \\ \bottomrule
 &  &  &  & 
\end{tabular}
 \caption{The results of OWQ and OPTQ with group-wise quantization. lamb. = lambada, winog. = winogrande.}
 \label{table:group}
\end{table}

\subsection{Comparison with Group-wise Quantization}

Applying uniform quantization at fine-grained granularity significantly reduces quantization error while introducing some storage overhead for quantization hyperparameters. OPTQ utilizes this approach by dividing row vectors into groups (e.g., group size of 128 or 1024) and applying uniform quantization independently with different configurations. This expansion can be applied orthogonally to OWQ, so we can combine it with OWQ to assess any improvements. Results in \Cref{table:group} show that the improvement from fine-grained quantization is negligible, as OWQ already substantially enhances the 3-bit model's quality. Moreover, compared to grouped OPTQ with 128 group size, 3.01-bit OWQ's storage overhead is only about 10\% of grouped OPTQ overhead while achieving comparable or better perplexity and zero-shot accuracy. Thus, OWQ is a superior solution to the grouping technique.

\begin{table}[t]
        \centering
        \small
        \begin{tabular}{ c | c @{\hskip 0.12in} c @{\hskip 0.12in} c @{\hskip 0.12in} c @{\hskip 0.12in} c }
            \toprule
            Equation & $\lambda_j || \Delta W_{:,j} ||_2^2$ & $\lambda_{j}$ & $|| \Delta W_{:,j} ||_2^2$ & $\Sigma |W_{:,j}|$ \\
            \midrule
            OPT-6.7B & \textbf{11.23} & 11.25 & 13.08 & 14.77 \\
            LLaMA 7B & \textbf{6.64} & 6.76 & 11.57 & 10.12\\
            \bottomrule
        \end{tabular}
        \caption{LLaMA WikiText-2 perplexity (lower is better).}
        \label{table:weak_column_selection_metrics}
\end{table}

\subsection{Weak Column Selection Metrics}
In this paper, we propose to use both the Hessian matrix and weight perturbations for weak column selection (\cref{eq: sensitivity}), to minimize layer output error. \Cref{table:weak_column_selection_metrics} additionally presents perplexity results for various weak column selection metrics. It is obvious that the results for the 1st ($\lambda_j || \Delta W_{:,j} ||_2^2$, our proposed selection metric) and 2nd ($\lambda_{j}$, Hessian only) columns are considerably better than those for the 3rd ($|| \Delta W_{:,j} ||_2^2$, weight error only).
This indicates that (1) minimizing layer-wise error is a valid objective for final accuracy, and (2) it is necessary to account for all factors contributing to output activation error, rather than simply focusing on minimizing weight error.

\begin{table}[t]
    \centering
    \small
    \begin{tabular}{ c | c @{\hskip 0.14in} c @{\hskip 0.14in} c @{\hskip 0.14in} c @{\hskip 0.14in} c @{\hskip 0.14in} c }
    \toprule
    Eff. bit & 3.005 & 3.01 & 3.05 & 3.10 & 3.20 & 3.30 \\
    \midrule
    Wiki-2 ($\downarrow$) & 11.38 & 11.22 & 11.19 & 11.17 & 11.15 & 11.13 \\ 
    Lat. (\%) & 2.89 & 2.91 & 4.19 & 6.21 & 6.86 & 8.69\\
    Mem. (\%) & 0.21 & 0.41 & 2.04 & 3.95 & 7.60 & 10.95 \\
    \bottomrule
   \end{tabular}
   \caption{Effective bit-width sweep using OPT-6.7B model. Lat. = Latency Overhead, Mem. = Memory Overhead.}
   \label{table:ratio_sweep}
\end{table}

\subsection{Varying Ratios of Weak Columns}
We report the trade-offs in perplexity, latency, and memory usage for varying ratios in \Cref{table:ratio_sweep}. These overhead results are compared to the OPTQ 3-bit baseline. 
With just 3.005 bits, the model already surpasses the OPTQ 3-bit performance with negligible overhead. However, as the bit width increases, the performance gain saturates while overhead keeps increasing. As activation outliers are limited to only a few dimensions, even a small number of weak columns can lead to noticeable performance improvements. In the main tables, we focused on two specific ratios, 3.01 and 3.1-bit, as they effectively illustrate the trend. 

\section{Conclusion}
The presence of activation outliers has been identified as a significant challenge in LLM activation quantization. We found that even in weight quantization, activation outliers can increase the sensitivity of certain weight columns, leading to significant quality degradation in a low-precision domain. To overcome this, we introduced a novel quantization scheme, OWQ. Compared to existing 3-bit quantization methods, OWQ improves quality notably with only negligible storage and computation overhead, sustaining the benefit of low-precision compression. In addition, we introduce the WCT scheme, which enables task-specific adaptation with minimal memory overhead and shows outstanding performance. We believe that our insights will promote the widespread adoption of LLMs.

\section{Acknowledgement}
This work was supported by IITP grant (MSIT, No.2021-0-00310) and NRF grant (MSIT, RS-2023-00213611, RS-2023-00228970) funded by the Korea government.

\bibliography{owq_arxiv24}

\newpage

%%%%%%%%%%%%%%%%%%%%%%%%%%%%%%%%%%%%%%%%%%
% APPENDIX
%%%%%%%%%%%%%%%%%%%%%%%%%%%%%%%%%%%%%%%%%%

\section{Appendix}
\section*{A\quad Main Proofs}
\subsection*{1\quad Proof of Eq. (5) in the Main Manuscript}
Our goal is to find the quantized weight $\hat{W}$ that minimizes the objective function:
\begin{equation}
\underset{{\hat{W}}}{\arg\min} \; E \; = \; \underset{{\hat{W}}}{\arg\min} \; || W X - \hat{W} X ||_2^2 \quad \text{s.t.} \; C(\hat{W}) < C_t \;. \\
\label{eq:error_a}
\end{equation}
we can reorganize the squared error term in \cref{eq:error_a} as the sum of squared errors corresponding to each output channel in the weight matrix:
\begin{equation}
E = \; \Sigma_{i=1}^{C_{out}} \; ||W_{i,:} X - \hat{W}_{i,:} X||^2_2 = \Sigma_{i=1}^{C_{out}} \; E_i \;.
\label{eq:error2}
\end{equation}
From this decomposition, we can see that the overall error can be separated into the sum of the errors from each output channel. Therefore, our goal of minimizing overall error $E$ can be thought of as minimizing the error of each output channel $E_i$.
The error that occurs when quantizing a network can be approximated by a Taylor series expansion for the model weights as follows:
\begin{equation}
E_i = \frac{\partial E_i}{\partial W_{i,:}} \, \Delta W_{i,:}^T + \frac{1}{2} \, \Delta W_{i,:} \, H^{(i)} \, \Delta W_{i,:}^T \;,
\label{eq:taylor1}
\end{equation}
where $\Delta W_{i,:} = W_{i,:} - \hat{W}_{i,:}$ is the perturbation of the $i$'th row of the weight and $H^{(i)} = H = \partial^2 E_i / \partial W_{i,:}^2$ is the Hessian matrix containing the second order derivatives for all weights in the weight row $W_{i,:}$. Since $E_i$ is a quadratic function, all terms above the third order become zero. For a network trained with a local minimum of error, the first-order (gradient) term can be ignored:
\begin{equation}
E_i \approx \Delta W_{i,:} \, H \, \Delta W_{i,:}^T  \; .
\label{eq:taylor2}
\end{equation}
Therefore, the quantization error for each output channel can be approximated as \cref{eq:taylor2}.

\begin{table*}[!htbp]
    \begin{center}
    \small
    \begin{tabular}{ c | c | l l l l l l l l }
        \toprule
        OPT & Bits & \quad 125M & 350M & 1.3B & 2.7B & 6.7B & 13B & 30B & 66B   \\
        \midrule
        full & 16 & \quad 38.99 & 31.08 & 20.29 & 17.97 & 15.77 & 14.52 & 14.04 & 13.36 \\
        \midrule
        RTN & 4 & \quad 53.89 & 36.79 & 57.30 & 31.05 & 18.84 & 16.51 & 15.40 & 225.66 \\
        OPTQ & 4 & \quad 45.66 & 33.35
 & 21.60	& 18.95	& 16.32	& 14.83	& 14.21	& 13.52
\\
        \textbf{OWQ} & 4.01 & \quad $\textbf{41.76}^{*}_{\pm.77}$ & $\textbf{32.36}^{*}_{\pm.08}$ & $\textbf{21.39}_{\pm.15}$ & $\textbf{18.34}_{\pm.06}$ & $\textbf{15.82}_{\pm.04}$ & $\textbf{14.77}_{\pm.01}$ & $\textbf{14.13}_{\pm.02}$ & $\textbf{13.42}_{\pm.01}$ \\
        \midrule
        OPTQ & 3 & \quad 133.78 & 44.36 & 30.61 & 24.95 & 19.21 & 16.40 & 15.11 & 14.33 \\
        \textbf{OWQ} & 3.01 & \quad $\textbf{49.29}^{*}_{\pm.29}$ & $\textbf{38.22}^{*}_{\pm.23}$ & $\textbf{23.64}_{\pm.19}$ & $\textbf{19.88}_{\pm.08}$ & $\textbf{16.33}_{\pm.07}$ & $17.00_{\pm.14}$ & $\textbf{14.36}_{\pm.04}$ & $\textbf{13.67}_{\pm.01}$ \\
        \textbf{OWQ} & 3.1 & \quad $\textbf{46.84}_{\pm.50}$ & $\textbf{37.36}_{\pm.16}$ & $\textbf{22.30}_{\pm.11}$ & $\textbf{19.33}_{\pm.04}$ & $\textbf{16.28}_{\pm.04}$ & $\textbf{14.99}_{\pm.03}$ & $\textbf{14.33}_{\pm.03}$ & $\textbf{13.60}_{\pm.04}$ \\
        \bottomrule
    \end{tabular}
    \end{center}
    \caption{OPT Penn Treebank (PTB) perplexity.}
    \label{table:opt_ptb}
\end{table*}

\begin{table*}[!htbp]
    \begin{center}
    \small
    \begin{tabular}{ c | c | l l l l l l l l }
        \toprule
        OPT & Bits & \quad 125M & 350M & 1.3B & 2.7B & 6.7B & 13B & 30B & 66B   \\
        \midrule
        full & 16 & \quad 26.56 & 22.59 & 16.07 & 14.34 & 12.71 & 12.06 & 11.44 & 10.99 \\
        \midrule
        RTN & 4 & \quad 33.91 & 26.21 & 24.51 & 18.43 & 14.36 & 13.36 & 13.46 & 309. \\
        OPTQ & 4 & \quad 29.42 & 24.14 & 16.73 & 14.85 & 12.99 & 12.24 & 11.56 & 11.08 \\
        \textbf{OWQ} & 4.01 & \quad $\textbf{27.93}^{*}_{\pm.01}$ & $\textbf{23.37}^{*}_{\pm.02}$ & $\textbf{16.49}_{\pm.00}$ & $\textbf{14.60}_{\pm.01}$ & $\textbf{12.83}_{\pm.00}$ & $\textbf{12.17}_{\pm.00}$ & $\textbf{11.49}_{\pm.00}$ & $\textbf{11.02}_{\pm.00}$ \\
        \midrule
        OPTQ & 3 & \quad 42.64 & 29.90 & 20.46 & 17.48 & 14.56 & 13.16 & 12.14 & 11.53 \\
        \textbf{OWQ} & 3.01 & \quad $\textbf{31.28}^{*}_{\pm.07}$ & $\textbf{26.40}^{*}_{\pm.12}$ & $\textbf{17.69}_{\pm.02}$ & $\textbf{15.36}_{\pm.01}$ & $\textbf{13.23}_{\pm.01}$ & $13.29_{\pm.04}$ & $\textbf{11.69}_{\pm.01}$ & $\textbf{11.17}_{\pm.00}$\\
        \textbf{OWQ} & 3.1 & \quad $\textbf{30.01}_{\pm.04}$ & $\textbf{25.95}_{\pm.07}$ & $\textbf{17.14}_{\pm.00}$ & $\textbf{15.16}_{\pm.01}$ & $\textbf{13.17}_{\pm.01}$ & $\textbf{12.42}_{\pm.01}$ & $\textbf{11.67}_{\pm.00}$ & $\textbf{11.16}_{\pm.00}$ \\
        \bottomrule
    \end{tabular}
    \end{center}
    \caption{OPT C4 perplexity.}
    \label{table:opt_c4}
\end{table*}

\begin{table}[!htbp]
        \centering
        \setlength\tabcolsep{3.4pt}
        \small
        \begin{tabular}{ c | c | l l l l }
            \toprule
            LLaMA & Bits & \quad 7B & 13B & 30B & 65B \\
            \midrule
            full & 16 & \quad 10.12 & 9.08 & 8.16 & 8.88 \\
            \midrule
            RTN & 4 & \quad 11.25 & 9.78 & 8.65 & 10.66 \\
            OPTQ & 4 & \quad 11.37 & 9.49 & 8.43 & 9.84 \\
            \textbf{OWQ} & 4.01 & \quad $\textbf{10.67}_{\pm.17}$ & $\textbf{9.30}_{\pm.04}$ & $\textbf{8.29}_{\pm.01}$ & $\textbf{9.56}_{\pm1.86}$ \\
            \midrule
            OPTQ & 3 & \quad 15.05 & 11.67 & 9.92 & 10.37 \\
            \textbf{OWQ} & 3.01 & \quad $\textbf{12.46}_{\pm.50}$ & $\textbf{10.02}_{\pm.04}$ & $\textbf{8.69}_{\pm.01}$ & $\textbf{8.33}_{\pm.06}$ \\
            \textbf{OWQ} & 3.1 & \quad $\textbf{11.14}_{\pm.08}$ & $\textbf{9.73}_{\pm.06}$ & $\textbf{8.62}_{\pm.02}$ & $\textbf{9.87}_{\pm.87}$ \\
            \bottomrule
        \end{tabular}
        \caption{LLaMA Penn Treebank (PTB) perplexity.}
        \label{table:llama_ptb}
\end{table}

\begin{table}[!htbp]
        \centering
        \setlength\tabcolsep{4pt}
        \small
        \begin{tabular}{ c | c | l l l l }
            \toprule
            LLaMA & Bits & \quad 7B & 13B & 30B & 65B \\
            \midrule
            full & 16 & \quad 7.34 & 6.80 & 6.13 & 5.98 \\
            \midrule
            RTN & 4 & \quad 8.12 & 7.23 & 6.54 & 6.45 \\
            OPTQ & 4 & \quad 7.77 & 7.07 & 6.39 & 6.31 \\
            \textbf{OWQ} & 4.01 & \quad $\textbf{7.67}_{\pm.01}$ & $\textbf{6.97}_{\pm.00}$ & $\textbf{6.25}_{\pm.00}$ & $\textbf{6.21}_{\pm.22}$ \\
            \midrule
            OPTQ & 3 & \quad 10.26 & 8.70 & 7.82 & 7.28 \\
            \textbf{OWQ} & 3.01 & \quad $\textbf{8.62}_{\pm.04}$ & $\textbf{7.43}_{\pm.00}$ & $\textbf{6.74}_{\pm.01}$ & $\textbf{6.31}_{\pm.00}$ \\
            \textbf{OWQ} & 3.1 & \quad $\textbf{8.16}_{\pm.03}$ & $\textbf{7.31}_{\pm.01}$ & $\textbf{6.57}_{\pm.00}$ & $\textbf{6.48}_{\pm.05}$ \\
            \bottomrule
        \end{tabular}
        \caption{LLaMA C4 perplexity.}
        \label{table:llama_c4}
\end{table}

\section*{B\quad Experiment Details}

The implementation of OWQ is based on OPTQ (GPTQ) official GitHub~\cite{gptq2022}.

\subsection*{1\quad Evaluation Settings}
We measured language modeling perplexity on datasets WikiText-2~\cite{merity2016pointer}, PTB~\cite{marcus1994penn}, and C4~\cite{raffel2020exploring}. The validation set is concatenated using two newlines as separators in WikiText-2 and a space as a separator in the PTB and C4 and then the concatenated data is tokenized using the default HuggingFace~\cite{wolf2019huggingface} tokenizer for each model.

\subsection*{2\quad Effective Bit-Width}
\bigbreak
In the main manuscript, we described that we store a complete low-precision matrix with zero-filled weak columns and additional fp16 weak columns (latency-favored method). Another possible option is storing the reduced size low-precision matrix to keep the total number of weight elements the same and to avoid storing unnecessary zeros corresponding to weak columns (storage-favored method). The latter method has trade-offs: it can save more storage, but it has more overhead in the actual operation.

We calculated the effective bit-width using the storage-favored method for the accuracy results in the main manuscript and the appendix. However, the storage overhead is similar for both methods as there are few weak columns. 

If we add the storage overhead of the zero-filled low-precision matrix to our 3.01-bit case, it becomes 3.012-bit, which is negligible overhead.

\section*{C\quad Additional Results}

\subsection*{1\quad Additional Perplexity Results}
Tables in this section (\Cref{table:opt_ptb}, \ref{table:opt_c4}, \ref{table:llama_ptb}, and \ref{table:llama_c4}) show additional language generation task results. 

In this section, results with * indicate that we used an extra 0.05 bits instead of 0.01 bits; there are few or no weak columns in the budget of 0.01 bits due to the small model dimension. Similar to the OPTQ, the calibration data were sampled from the C4 training set, so measuring perplexity on C4 is not a fully zero-shot situation. Please note that we additionally applied the true-sequential (TS) and act-order (AO) options for OPTQ results, while the proposed OWQ doesn't use them. TS and AO options were originally not introduced in OPTQ paper, however, they can boost the OPTQ results in certain models.

\begin{table*}[t]
    \centering
    \small
    \setlength\tabcolsep{4pt}
    \begin{tabular}{ c | c | c | c | c | c }
    \toprule
    layers & $W_{k}, W_{q}$ & $W_{k}, W_{v}$ & $W_{k}, W_{fc1}$ & $W_{q}, W_{v}$ & $W_{v}, W_{fc1}$ \\
    \midrule
    WikiText-2 & \textbf{11.26} & 19.87 & 22.25 & 111.57 & 89.92 \\
    \midrule
    layers & $W_{k}, W_{q}, W_{v}$ & $W_{k}, W_{q}, W_{fc1}$ & $W_{k}, W_{q}, W_{out}$ & $W_{q}, W_{v}, W_{out}$ \\
    \midrule
    WikiText-2 & 11.31 & \textbf{11.24} & 11.26 & 111.81 \\
    \bottomrule
   \end{tabular}
   \caption{Layer-wise sensitivity analysis on OPT-6.7B model.}
   \label{table:layer_sweep}
\end{table*}

\subsection*{2\quad Layer-wise Quantization Sensitivity}

In our experiments, we uniformly allocated extra storage budget to linear layers but observed varying sensitivity of weak columns across layers. Using the OPT-6.7B model, we assessed layer-by-layer sensitivity by applying OWQ to specific layers within a fixed capacity budget. Our results in \Cref{table:layer_sweep} indicate the best accuracy is achieved when weak columns are preserved simultaneously from both key and query weights. Performance suffers significantly if both layers aren't addressed together, especially when we only apply OWQ to queries. OWQ's quantization configuration search stage is more sensitive to weak columns in key and query layers with higher magnitudes than other layers. Improvements could be made by allocating budgets at different rates according to each layer's sensitivity. However, the rate tuning is prohibitively expensive, making implementing this approach challenging. Developing a method to tune layer-wise weak column numbers is a potential future research direction.

\subsection*{3\quad Generation Examples}

Below paragraphs are examples of questions from the Vicuna Benchmark \cite{vicuna2023} we used for our evaluation and corresponding answers.

\begin{quote}
\textbf{User}: What are the most effective ways to deal with stress?

: \texttt{1. Meditation. Studies have shown that meditation can help relieve stress, anxiety, and depression.
2. Exercise. Exercising can release endorphins, which are chemicals in the brain that can improve mood.
3. Deep breathing. Deep breathing exercises can help calm your mind and body.
4. Sleep. Getting enough sleep can help your body and mind recover from the stresses of the day.
5. Healthy eating. Eating a healthy diet can help provide your body with the nutrients it needs to function properly.}
\end{quote}

\begin{quote}
\textbf{User}: How can I develop my critical thinking skills?

: \texttt{1. Read widely. Seek out a variety of viewpoints, including those that challenge your own.
2. Question assumptions. Ask yourself whether you're taking things at face value or critically evaluating evidence and arguments.
3. Avoid confirmation bias. Don't seek out information that confirms your existing beliefs. Instead, look for evidence that challenges them.
4. Don't be afraid to change your mind. Be open to new information and be willing to update your beliefs in light of it.
5. Engage in debate. Participate in discussions where you can test your ideas against those of others..}
\end{quote}

\section*{D\quad Calculation Details}

\subsection*{1\quad Cost comparison between OWQ and group-wise quantization}
For the OPT model family, the memory overhead from full-precision values for both OWQ and Grouping can be determined using the following equations:

\begin{equation}
O_{OWQ} = (12d^2) \cdot \frac{(t-w)}{(16-w)} + 9d, 
\label{eq:owq_overhead}
\end{equation}

\begin{equation}
O_{group} = \frac{12d^2}{g}.
\label{eq:group_overhead}
\end{equation}

In these formulas, $d$ symbolizes the model's hidden size, $w$ is quantization bit, $t$ is the average target bit-width, and $g$ stands for the group size. Based on these computations, OWQ with a target bit-width of 3.01 has an overhead roughly equivalent to that of a group size of 1024, and OWQ with a target bit-width of 3.1 closely matches the overhead of a group size of 128.

However, it is crucial to highlight that OWQ and group-wise quantization are founded on different principles (preserving weak parts of weights vs. fine-grained quantization), allowing for orthogonal application of group-wise quantization to OWQ, and providing further optimization possibilities. In the case of the LLaMA family, $+ 9d$ in \cref{eq:owq_overhead} is slightly different ($+ 10.3d$), but the overall cost is almost the same.

\subsection*{2\quad Determination of \( k \) Value}

Consider OPT model with a hidden size denoted as \( d \).
There are \( 12d^2 \) elements in a single OPT decoder block, and since not all layer weights have the same column size, we make the single decoder block have \( rd^2 \) fp16 elements and distribute them evenly to each layer.

When undergoing quantization to \( w \)-bits, the average target bit-width ($t$) can be derived as:
\[
t = \frac{w \cdot (12d^2 - rd^2) + 16 \cdot rd^2}{12d^2} = w + \frac{(16-w)}{12} \cdot r, 
\]
and consequently,
\[
r = \frac{(t-w)}{16-w} \cdot 12.
\]

This r value is then uniformly distributed among all layers in a given block. This means every layer in self-attention will consist of \( k \) weak columns, defined as:
\[
k = round\left(\frac{r}{\# \; of \; layers} \cdot d\right).
\]

When working with feed-forward layers, \( k \) is adjusted to ensure its overhead aligns with the self-attention layer's overhead. The overarching goal is to balance the number of elements of weights retained as fp16 within each layer.

\subsection*{3 \quad Comparison of the number of Learnable Parameters between WCT and QLoRA}
We formulate and compare the number of learnable parameters of WCT and QLoRA for the LLaMA model.
Since the LLaMA model is a transformer-based structure in which the same decoder block is stacked repeatedly, the number of parameters in a single decoder block was calculated in bits for accurate calculation.
In common, $d$ is the hidden state size of the model and $D$ is the intermediate size of the model in the feed-forward network. 
The number of learnable parameters in WCT and QLoRA is as follows: 
\begin{equation}
WCT : QLoRA = (5d + 2D) k : (11d + 3D) r, 
\end{equation}
where $r$ is the rank of the adapter modules and $k$ is the number of weak columns. 
For the LLaMA 7B model with $d = 4096$ and $D = 11008$ with $r = 64$ and $k = 8$ case, WCT has only 6.8\% learnable parameters compared to QLoRA.

\end{document}